\documentclass[11pt]{article}

\usepackage[preprint]{acl}

\usepackage{times}
\usepackage{latexsym}
\usepackage{pifont}

\usepackage[T1]{fontenc}

\usepackage[utf8]{inputenc}

\usepackage{microtype}

\usepackage{inconsolata}

\usepackage{graphicx}

\usepackage{enumitem}
\usepackage{booktabs}
\usepackage{multirow}
\usepackage[table]{xcolor}

\definecolor{mygray}{gray}{0.9} 
\definecolor{myblue}{HTML}{ECF4FF}
\usepackage{amsmath}
\usepackage{tcolorbox}
\usepackage{stfloats}
\usepackage{algorithm}
\usepackage{algorithmic}

\title{MemoSight: Unifying Context Compression and Multi-Token Prediction for Reasoning Acceleration}

\author{
  \textbf{Xinyu Liu\textsuperscript{1}},
  \textbf{Xin Liu\textsuperscript{1}},
  \textbf{Bo Jin\textsuperscript{1}},
  \textbf{Runsong Zhao\textsuperscript{1}},
  \textbf{Pengcheng Huang\textsuperscript{1}},
  \\
  \textbf{Junhao Ruan\textsuperscript{1}},
  \textbf{Bei Li\textsuperscript{2}},
  \textbf{Chunyang Xiao},
  \textbf{Chenglong Wang\textsuperscript{1}}
  \textbf{Tong Xiao\textsuperscript{1,3}}
  \textbf{Jingbo Zhu\textsuperscript{1,3}}
  \\
  \textsuperscript{1} \normalsize{School of Computer Science and Engineering, Northeastern University, China} \\
  \textsuperscript{2} \normalsize{Meituan Inc.}
  \textsuperscript{3} \normalsize{NiuTrans Research, Shenyang, China} \\
  \normalsize{lxy1051493182@gmail.com} \\
  \normalsize{\{xiaotong, zhujingbo\}@mail.neu.edu.com}
}

\begin{document}
\maketitle


\begin{abstract}

While chain-of-thought (CoT) reasoning enables LLMs to solve challenging reasoning tasks, the linear growth of the KV cache leads to substantial memory and inference overhead. 
Existing approaches such as context compression and multi-token prediction (MTP) improve efficiency from two complementary directions by compressing historical tokens and generating future tokens in parallel. 
However, effectively combining them remains challenging due to their different training paradigms and architectural assumptions.
In this work, we propose MemoSight (\textbf{Memo}ry-Fore\textbf{sight}-Based Reasoning), a unified framework that integrates context compression and MTP to improve inference efficiency while preserving CoT performance. 
MemoSight adopts a shared minimalist design based on special tokens and token-specific positional layouts for both compression and parallel prediction. 
Experiments on four reasoning benchmarks show that, compared to the vanilla SFT baseline, MemoSight reduces KV cache usage by up to 66\% and improves inference speed by 56\%, while incurring less than a 3\% drop in average reasoning accuracy, yielding a better efficiency--accuracy trade-off than existing CoT compression methods.

\end{abstract}

    

\begin{figure*}[h]
    \centering
    \includegraphics[width=1.0\linewidth]{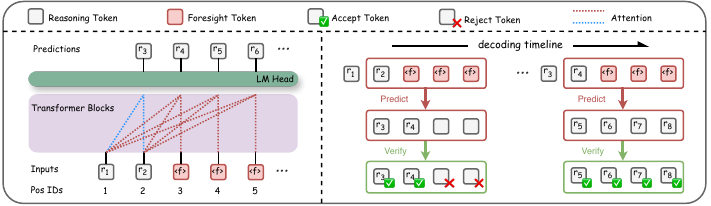}
    \caption{Illustration of \textbf{foresight-token-based acceleration} ($d=3$). \textbf{(Left)} Foresight tokens $\langle \text{f} \rangle$ are inserted after the current reasoning prefix with increasing position IDs; each foresight token attends to reasoning tokens and itself, predicting a future reasoning token through the shared LM head. \textbf{(Right)} During decoding, the model first generates multiple candidate future tokens in a single forward pass (\textit{Predict}), then verifies them in parallel (\textit{Verify}). Correct predictions are accepted and appended to the reasoning sequence, while incorrect predictions are rejected.}
    \label{fig:foresight_infer}
    \vspace{-5pt}
\end{figure*}

\section{Introduction}



Long Chain-of-Thought (CoT) reasoning has become an important mechanism for improving the abilities of large language models (LLMs)~\cite{zhao2023survey, Guo_2025, jaech2024openai}. However, in Transformer-based architectures~\cite{vaswani2017attention}, the Key-Value (KV) cache grows linearly with the number of generated tokens, creating latency and memory bottlenecks that hinder the real-world deployment of long CoT reasoning~\cite{arora2025training}.

Context compression and multi-token prediction (MTP) address these inference bottlenecks from complementary aspects. \textbf{Context compression}~\cite{chang2024efficient, zhang2025lightthinker} condenses reasoning traces into compact ``memory tokens'', allowing the model to represent past reasoning states with a much shorter context during inference. This reduces KV cache usage and lowers decoding cost. \textbf{Multi-token prediction (MTP)}~\cite{gloeckle2024better, liu2024deepseek}, in contrast, trains models to predict multiple future tokens at each step rather than only the next token. When combined with speculative decoding~\cite{cai2024medusa,li2024eagle}, MTP can further speed up inference. In short, context compression reduces the cost of processing \emph{past} reasoning traces, while MTP accelerates \emph{future} token generation. Their combination offers the potential for further inference efficiency gains.

However, we find that a naive combination of MTP and context compression degrades performance, as shown in Section~\ref{subsec:MTP}. 
Specifically, we follow the MTP design of DeepSeek-V3~\cite{liu2024deepseek} by augmenting the model with additional Transformer blocks and fine-tuning the resulting model on the same context compression data. 
We hypothesize that this failure arises from a mismatch between conventional MTP designs and the context compression setting. 
Existing MTP methods are mainly developed for large-scale pre-training and introduce substantial architectural changes with many newly added parameters. 
In contrast, context compression is learned during supervised fine-tuning through lightweight adaptation. 
As a result, conventional MTP designs are not well aligned with the SFT-based context compression objective, leading to suboptimal performance.

In this work, we integrate context compression and MTP in our proposed MemoSight (\textbf{Memo}ry-Fore\textbf{sight}-Based Reasoning) framework. MemoSight requires no modifications to LLM architectures and provides a special-token-based unified framework through the following design:

\begin{itemize}[topsep=4pt, itemsep=4pt, parsep=2pt, leftmargin=*]
\item \textbf{Special Tokens as Carriers.} Inspired by recent developments from context compression~\cite{zhang2025lightthinker} and MTP~\cite{gerontopoulos2025multi}, we incorporate both mechanisms into our framework by leveraging special tokens with different roles: memory tokens condense reasoning steps into compact representations; foresight tokens trigger multi-token prediction. 
\item \textbf{Position-Aware Alignment.} 
Prior work has shown effectiveness by designing position layout for context compression~\cite{zhao2024position} and MTP~\cite{gerontopoulos2025multi} in isolation. MemoSight introduces a tailored position layout for each type of special tokens, aligning memory tokens with past reasoning and foresight tokens with future predictions. 
\end{itemize}


By integrating context compression and MTP within a unified framework, MemoSight combines the benefits of both methods. As we demonstrate in Section~\ref{sec:experiments}, evaluations on four reasoning benchmarks show that, compared to the vanilla CoT fine-tuning baseline, MemoSight reduces KV cache usage by 66\% and improves inference speed by 56\%, with only a 3\% average accuracy drop. Compared to the context compression baseline LightThinker, MemoSight improves average accuracy by 3 points and inference speed by 27\%, yielding a better efficiency--performance tradeoff.

\section{MemoSight}
\label{sec:method}


This section presents MemoSight (Memory-Foresight-Based Reasoning), a framework that jointly models context compression and multi-token prediction (MTP) for efficient CoT reasoning. Notably, MemoSight employs unified techniques through special tokens and dedicated positional layout for both context compression and MTP, requiring no architectural modifications; such unified and minimalist design enables MemoSight to achieve benefits from both techniques through supervised fine-tuning. In this section, we first introduce the two inference accelerations (Section~\ref{sec:inference_pipeline}) MemoSight incorporates, then how we train MemoSight to enable these two accelerations: Section~\ref{sec:data_construction} describes the training sequence construction, followed by the training framework with customized attention (Section~\ref{sec:training}).

\subsection{Iterative Inference Pipeline}
\label{sec:inference_pipeline}

Given a prompt $P = [p_1, \dots, p_{|P|}]$, a reasoning LLM parameterized by $\theta$ autoregressively generates a reasoning sequence $R = [r_1, \dots, r_{|R|}]$ according to
$p_\theta(r_t \mid P, r_{<t})$. MemoSight modifies this decoding process by alternating between accelerated generation and context compression. To support these inference modes, MemoSight introduces two types of special tokens~\footnote{Throughout the paper, we use $\langle \cdot \rangle$ to denote special tokens introduced for modeling purposes.}: \emph{foresight tokens $\langle \text{f} \rangle$} for MTP and \emph{memory tokens $\langle \text{m} \rangle$} for context compression.

\begin{figure*}[h]
    \centering
    \includegraphics[width=1.0\linewidth]{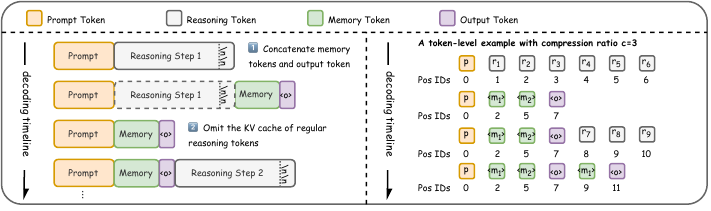}
    \caption{Illustration of \textbf{memory-token-based compression}. 
    \textbf{(Left)} After each reasoning step, the model compresses the reasoning trace into memory tokens and discards the verbose KV cache. Future reasoning attend only to the prompt, memory tokens, and output tokens.
    \textbf{(Right)} A token-level example with compression ratio $c=3$. Memory token count scales with the reasoning step length (AMA), and their position IDs are uniformly interpolated over the compressed span (UPL).}
    \label{fig:memory_infer}
    \vspace{-5pt}
\end{figure*}

\subsubsection{Foresight Token-Based Acceleration}

Formally, MTP predicts multiple future tokens according to $p_\theta(r_{t:t+d} \mid P,r_{<t})$, where $d$ is the MTP horizon and corresponds to the number of tokens predicted beyond the next token. Conventional methods achieve this by branching the current hidden states into $d$ parallel~\cite{gloeckle2024better} or sequential~\cite{liu2024deepseek} prediction heads, each responsible for predicting a different future offset.

Motivated by~\cite{gerontopoulos2025multi}, MemoSight instead performs MTP using special tokens without modifying the model architecture. We append $d$ foresight tokens $\langle \text{f} \rangle$ after the current position and model
\begin{equation}
    p_\theta\Big(r_{t:t+d} \;\Big|\; P,r_{<t}, \underbrace{\langle \text{f} \rangle, \dots, \langle \text{f} \rangle}_{d}\Big).
\end{equation}
As illustrated in Figure~\ref{fig:foresight_infer}(left), each foresight token corresponds to a different future offset and is assigned a sequentially increasing position ID, indicating which future token it predicts (e.g., $\langle \text{f} \rangle$ with position ID 3 predicts the token $r_4$ at position 4). Each foresight token attends only to preceding regular tokens and itself, preventing information leakage across future offsets. As a result, a single forward pass produces $d+1$ draft tokens, including the standard next-token prediction from the current token representation.

As shown in Figure~\ref{fig:foresight_infer}(right), during inference, we append $d$ foresight tokens after the current position to generate $d+1$ draft tokens in a single forward pass (with $d=3$ in the figure). We then perform a parallel verification pass over all draft tokens, keep the accepted tokens~\footnote{For example, we accept the prefix of tokens, each of which coincide with argmax prediction.}, and continue generation from the last accepted position. For example, in Figure~\ref{fig:foresight_infer}(right), $r_3, r_4$ are accepted in the first MTP, which are used in the following predcitions.

\subsubsection{Memory-Token-Based Compression}
\label{sec:memory_compression}

Following prior work~\cite{zhang2025lightthinker}, MemoSight partitions the reasoning sequence $R=[r_1,\dots,r_{|R|}]$ into a sequence of reasoning steps $R_1,\dots,R_n$ using a delimiter.~\footnote{Following~\cite{zhang2025lightthinker}, we use ``\texttt{\textbackslash n\textbackslash n}'' as our delimiter.} Once the model generates the step delimiter token, it appends both memory tokens and a dedicated output token $\langle \text{o} \rangle$ after the current step $R_i$. The memory tokens compress $R_i$ into compact representations, while $\langle \text{o} \rangle$ enables continued generation over the compressed context. The regular reasoning tokens in $R_i$ are then removed from the KV cache to reduce the effective context length.

We illustrate this memory token based compression process in of Figure~\ref{fig:memory_infer} (left); on top of this, MemoSight further introduces two improvements to enhance the compression effectiveness that we illustrate in Figure~\ref{fig:memory_infer} (right):

\noindent \textbf{Adaptive Memory Allocation (AMA)}. Prior methods assign a fixed number of memory tokens to each reasoning step. However, reasoning steps can vary substantially in length and information density. Instead, MemoSight allocates memory tokens according to a predefined compression ratio $c$. For each reasoning step $R_i$, we compute the number of memory tokens as $l=\lceil |R_i|/c \rceil$ and append memory tokens $M_i=[m_1,\dots,m_l]$ after $R_i$. AMA thus enables more flexible compression for long reasoning traces.

\noindent \textbf{Uniform Position Layout (UPL)}. Prior methods assign consecutive new position IDs to memory tokens. After reasoning tokens are evicted, the remaining memory-token positions become sparsely distributed with large positional gaps, disrupting relative positional relationships in attention mechanisms such as RoPE~\cite{su2024roformer}. Motivated by~\citet{zhao2024position}, MemoSight instead assigns memory-token position IDs by uniformly interpolating across the positional span of each reasoning step. Specifically, we divide the current reasoning step $R_i$ into $l=\lceil |R_i|/c \rceil$ sequential groups,\footnote{Given compression ratio $c$, each group contains $c$ tokens except possibly the last group.} and assign each memory token the center position of its corresponding group. This preserves a uniform positional distribution of memory tokens across the historical context. Moreover, UPL places memory tokens closer to their corresponding content in positional space, facilitating better compression.


\subsubsection{MemoSight Inference}

During inference, MemoSight alternates between foresight-token-based acceleration and memory-token-based compression~\footnote{We provide a detailed algorithmic description of this inference pipeline in Appendix~\ref{appendix:infer_procedure}.}. Within each reasoning step $R_i$, the model performs parallel future-token prediction for accelerated decoding. Once $R_i$ is completed, the model compresses it into memory tokens $M_i$ and continues generation over the compressed context. These two inference modes are complementary: foresight-token acceleration reduces the number of decoding iterations through parallel generation, while memory-token-based compression controls KV cache growth during long reasoning processes. When MTP and compression are well trained, their combination is expected to improve both inference speed and memory efficiency for CoT reasoning.

\subsection{Data Construction Pipeline}
\label{sec:data_construction}

To train MemoSight to perform context compression and MTP, we construct training instances from standard CoT trajectories by inserting memory tokens $\langle \text{m} \rangle$ and foresight tokens $\langle \text{f} \rangle$ with customized position IDs.

\paragraph{CoT Augmentation.}

Given a CoT trajectory $R = [R_1, \dots, R_n]$, where each reasoning step is represented as $R_i = [r_1^i, r_2^i, \dots, r_{|R_i|}^i]$, we first augment each reasoning step with foresight tokens. Specifically, for each token $r_t^i$, we insert a foresight token $\langle \text{f} \rangle$ immediately before it, yielding
\begin{equation*}
\tilde{R}_i = [\langle \text{f} \rangle, r_1^i, \langle \text{f} \rangle, r_2^i,
\dots,
\langle \text{f} \rangle, r_{|R_i|}^i].
\end{equation*}
During training, each foresight token is optimized to predict the token $d'$ steps ahead, where $d' \in \{0,\dots,d\}$ is randomly sampled for each training sample~\footnote{When $d'=0$, the foresight token reduces to standard next-token prediction.}. This enables the model to learn future token prediction under different offsets.

For each reasoning step except the final one, we further append memory tokens $M_i = [\langle \text{m}_1 \rangle, \dots, \langle \text{m}_l \rangle]$ after $\tilde{R}_i$, where $l = \lceil |R_i|/c \rceil$ is determined by the compression ratio $c$, followed by an output token $\langle \text{o} \rangle$ marking the transition to the next reasoning step. Combined with the prompt $P$, the final training sequence is constructed as
\begin{equation*}
\mathcal{S} = [P, \tilde{R}_1, M_1, \langle \text{o} \rangle, \tilde{R}_2, M_2, \langle \text{o} \rangle, \dots, \tilde{R}_n].
\end{equation*}

\paragraph{Label Assignment.}

\begin{figure}[t]
    \centering
    \includegraphics[width=1.0\linewidth]{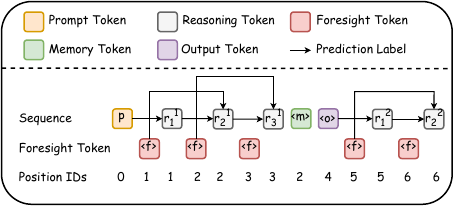}
    \caption{An example of sequence $\mathcal{S}$ with a compression ratio $c=3$ and foresight offset $d'=1$. The reasoning tokens ($r_1^1, r_2^1, r_3^1$) are compressed into the memory token $\langle \text{m} \rangle$ with an interpolated position ID (i.e., 2). An output token $\langle \text{o} \rangle$ then bridges the compressed context to the subsequent step. Foresight tokens $\langle \text{f} \rangle$ are inserted before $r_t^i$ with position IDs $t+d'-1$. The solid arrows indicate the prediction labels for each token type.}
    \label{fig:data_sample}
    \vspace{-5pt}
\end{figure}

We define supervision targets for different token types in the sequence $\mathcal{S}$. 
As illustrated in Figure~\ref{fig:data_sample}, excluding the prompt $P$, $\mathcal{S}$ contains four token types: reasoning, foresight, output, and memory tokens.

Reasoning tokens $r_t^i$ follow standard next-token prediction within each reasoning step $\tilde{R}_i$. No prediction target is assigned to the last reasoning token in $\tilde{R}_i$ (i.e., the step delimiter token ``\texttt{\textbackslash n\textbackslash n}'').

For each foresight token $\langle \text{f} \rangle$ preceding $r_t^i$, the prediction target is defined as $r_{t+d'}^i$. If $t+d' > |R_i|$, no prediction target is assigned.

For each output token $\langle \text{o} \rangle$, the prediction target is defined as the first token of the next reasoning step, i.e., $r_1^{i+1}$.
No prediction target is assigned to memory tokens $\langle \text{m} \rangle$; memory token parameters are learned solely through back propagations of other tokens.

\paragraph{Position ID Assignment.}

We assign dedicated position IDs (PIDs) to different token types in the sequence $\mathcal{S}$. Let $\rho_t^i$ denote the PID of reasoning token $r_t^i$. The output token $\langle \text{o} \rangle$ after reasoning step $\tilde{R}_{i}$ is assigned PID $\rho_{|R_i|}^i + 1$, such that reasoning and output tokens occupy consecutive positions in the augmented CoT sequence. As shown in Figure~\ref{fig:data_sample}, the $\langle \text{o} \rangle$ is assigned PID 4 following $r_3^1$. 

\begin{itemize}[topsep=4pt, itemsep=4pt, parsep=2pt, leftmargin=*]


\item \textbf{Memory Tokens.}
For memory tokens $M_i = [\langle \text{m}_1 \rangle, \dots, \langle \text{m}_l \rangle]$, we uniformly interpolate their PIDs over the positional span of the reasoning tokens in step $\tilde{R}_i$ to align with the UPL formulation in Section~\ref{sec:memory_compression}. For example, in Figure~\ref{fig:data_sample}, reasoning step $\tilde{R}_1$ spans PIDs 1--3. With compression ratio $c=3$, the memory token is assigned an interpolated PID 2.

\item \textbf{Foresight Tokens.}
For the foresight token $\langle \text{f} \rangle$ inserted before $r_t^i$, its PID is assigned as $\rho_{t+d'-1}^i$, i.e., the PID immediately preceding the prediction target $r_{t+d'}^i$ in $\tilde{R}_i$. This positional alignment encourages $\langle \text{f} \rangle$ to predict $r_{t+d'}^i$. For example, when $d'=1$, the foresight token before $r_1^1$ shares the same PID as $r_1^1$, encouraging it to predict $r_2^1$.

\end{itemize}

\subsection{Joint Training Framework}
\label{sec:training}


\begin{figure}[h]
    \centering
    \includegraphics[width=1.0\linewidth]{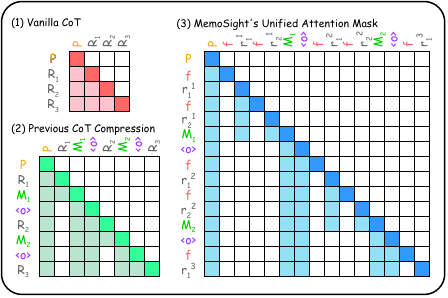}
    \caption{Comparison of training attention masks. (1) \textbf{Vanilla CoT} uses a standard causal mask. (2) In \textbf{LightThinker}, the attention restricts to memory tokens $\langle \text{m} \rangle$ and current reasoning tokens, discarding historical reasoning steps. (3) \textbf{MemoSight} follows LightThinker to discard historical reasoning steps in its attention but further customizes the attention patterns due to the foresight token introduction.}
    \label{fig:attn}
\vspace{-5pt}
\end{figure}


\paragraph{Attention Mask Strategy.}

A critical component of our training framework is the specialized attention mask illustrated in Figure~\ref{fig:attn}. Our attention design builds upon LightThinker while introducing additional masking rules for foresight tokens.

\begin{itemize}[topsep=4pt, itemsep=4pt, parsep=2pt, leftmargin=*]

\item \textbf{Compression Attention.}
For the current reasoning step $\tilde{R}_i$, standard reasoning tokens $r_t^i$ and memory tokens $M_i$ follow standard causal attention over preceding reasoning tokens. For previous reasoning steps, they attend exclusively to the prompt $P$, condensed memory tokens $M_{<i}$, and output tokens, while the verbose reasoning traces $\tilde{R}_{<i}$ are masked out. This design compels the model to rely entirely on compressed memory for historical context.


\item \textbf{MTP Attention.}
A foresight token $\langle \text{f} \rangle$ preceding $r_t^i$ shares an identical attention pattern to $r_t^i$, attending to $P$, $M_{<i}$, $\langle \text{o} \rangle$, the prefix reasoning tokens in the current step $\tilde{R}_i$, and itself. Its hidden state is masked from all future tokens, enabling foresight tokens to function as parallel predictive branches without interfering with the main reasoning process.

\end{itemize}


\paragraph{Joint Optimization Objective.}
A standard next-token prediction loss, $\mathcal{L}_{\text{NTP}}$, is applied to reasoning tokens and output tokens $\langle \text{o} \rangle$, while the loss on memory tokens $\langle \text{m} \rangle$ is masked out. The multi-token prediction loss, $\mathcal{L}_{\text{MTP}}$, is applied to foresight tokens $\langle \text{f} \rangle$ to predict their future counterparts. The overall training objective is:
\begin{equation*}
\mathcal{L}_{\text{MemoSight}} = \lambda \mathcal{L}_{\text{NTP}} + (1-\lambda) \mathcal{L}_{\text{MTP}},
\end{equation*}
where $\lambda$ controls the trade-off between standard next-token prediction and foresight prediction.

\begin{table*}[!t]
\centering
\tiny
\setlength{\tabcolsep}{1.5pt} 
\resizebox{\linewidth}{!}{

    \begin{tabular}{l ccc | ccc | ccc | ccc | ccc}
    \toprule
    
    \multirow{4}{*}{\textbf{Method}} & \multicolumn{3}{c}{\textbf{GSM8K}} & \multicolumn{3}{c}{\textbf{MMLU}} & \multicolumn{3}{c}{\textbf{GPQA}} & \multicolumn{3}{c}{\textbf{BBH}} & \multicolumn{3}{c}{\textbf{AVG.}} \\ 
    \cmidrule(lr){2-4} \cmidrule(lr){5-7} \cmidrule(lr){8-10} \cmidrule(lr){11-13} \cmidrule(lr){14-16}
    
    & Acc~$\uparrow$ & Speed~$\uparrow$ & Peak~$\downarrow$ 
    & Acc~$\uparrow$ & Speed~$\uparrow$ & Peak~$\downarrow$ 
    & Acc~$\uparrow$ & Speed~$\uparrow$ & Peak~$\downarrow$ 
    & Acc~$\uparrow$ & Speed~$\uparrow$ & Peak~$\downarrow$ 
    & Acc~$\uparrow$ & Speed~$\uparrow$ & Peak~$\downarrow$ \\ 
    \cmidrule{1-16}

    \rowcolor{mygray} \multicolumn{16}{c}{\textit{Qwen2.5-7B Series}} \\ 
    \cmidrule{1-16}
    CoT &{88.32}&{24.63}&{519} &{70.01}&{23.78}&{676} &{28.28}&{25.50}&{998} &{71.31}&{27.47}&{582} &{64.48}&{25.35}&{694} \\
    Distill-R1 &{60.58}&{24.82}&{522} &{31.26}&{21.49}&{1289} &{22.22}&{19.21}&{4383} &{51.31}&{29.35}&{905} &{41.34}&{23.72}&{1775} \\ 
    \cmidrule{1-16}
    Vanilla &{90.83}&{20.69}&{1559} &{65.92}&{18.6}&{2241} &{37.37}&{15.10}&{7184} &{81.01}&{24.32}&{2212} &{68.78}&{19.68}&{3299} \\
    \rowcolor{myblue} 
    \quad + H2O &{\underline{91.05}}&{13.85}&{1024} &\underline{62.32}&{10.36}&{1024} &{20.20}&{9.96}&\textbf{1024} &\textbf{75.76}&{14.14}&{1024} &{62.33}&{12.08}&\textbf{1024} \\ 
    \rowcolor{myblue} 
    \quad + SepLLM &{\textbf{91.28}}&{11.03}&{1024} &{59.20}&{10.97}&{1024} &{12.63}&{11.30}&\textbf{1024} &{69.70}&{14.27}&{1024} &{58.20}&{11.89}&\textbf{1024} \\ 
    \rowcolor{myblue} 
    LightThinker &{87.26}&\underline{22.56}&\textbf{684} &{60.66}&{\underline{22.58}}&\textbf{827} &\underline{37.37}&{\underline{22.38}}&\underline{1969} &{66.46}&{28.13}&\textbf{980} &\underline{62.94}&{\underline{23.91}}&\underline{1115} \\ 
    \cmidrule{1-16}
    \rowcolor{myblue} 
    \text{MemoSight} &{89.84}&\textbf{30.94}&\underline{745} &\textbf{63.49}&\textbf{27.54}&\underline{900} &\textbf{40.91}&\textbf{27.64}&{2102} &\underline{73.13}&\textbf{32.26}&\underline{1059} &\textbf{66.84}&\textbf{29.60}&{1202} \\ 
    \cmidrule{1-16}
    \rowcolor{mygray} \multicolumn{16}{c}{\textit{Llama3.1-8B Series}} \\ 
    \cmidrule{1-16}
    CoT &{64.52}&{21.41}&{497} &{60.18}&{19.23}&{698} &{24.75}&{21.40}&{2230} &{53.13}&{26.28}&{596} &{50.65}&{22.08}&{1005} \\ 
    Distill-R1 &{57.62}&{22.78}&{524} &{20.93}&{17.89}&{1611} &{30.81}&{16.40}&{5640} &{29.49}&{27.65}&{1146} &{34.71}&{21.18}&{2230} \\  
    \cmidrule{1-16}
    Vanilla &{89.31}&{18.33}&{1702} &{71.08}&{16.26}&{2800} &{37.37}&{14.17}&{6824} &{78.18}&{21.63}&{2515} &{68.99}&{17.60}&{3460} \\ 
    \rowcolor{myblue} 
    \quad + H2O &\textbf{89.31}&{11.65}&{1024} &\textbf{69.52}&{9.03}&{1024} &{23.23}&{8.51}&\textbf{1024} &\textbf{81.62}&{13.04}&{1024} &\underline{65.92}&{10.56}&\textbf{1024} \\ 
    \rowcolor{myblue} 
    \quad + SepLLM &\underline{88.32}&{19.74}&{1024} &\underline{64.85}&{19.02}&{1024} &{15.66}&{19.27}&\textbf{1024} &{72.32}&{24.12}&{1024} &{60.29}&{20.54}&\textbf{1024} \\ 
    \rowcolor{myblue} 
    LightThinker &{85.82}&\underline{21.38}&\textbf{664} &{60.86}&\underline{20.53}&\textbf{932} &\textbf{37.37}&{19.89}&\underline{1933} &{72.93}&\underline{26.11}&\underline{999} &{64.25}&\underline{21.98}&{1132} \\  
    \cmidrule{1-16}
    \rowcolor{myblue} 
    \text{MemoSight} &{87.19}&{\textbf{27.91}}&\underline{750} &{64.07}&{\textbf{25.89}}&\underline{933} &\underline{35.86}&\textbf{26.67}&{1942} &\underline{78.38}&\textbf{33.45}&\textbf{866} &\textbf{66.38}&\textbf{28.48}&\underline{1123} \\

    \bottomrule
    \end{tabular}
}

\caption{Main results on the Qwen2.5-7B and Llama3.1-8B models. Acceleration methods are highlighted in blue, with \textbf{bold} and \underline{underlined} values indicating the best and second-best results among them. Peak and Speed denote maximum context token counts and number of tokens generated per second, respectively.}
\label{table:exp_main}
\vspace{-5pt}
\end{table*}

\section{Experiments}
\label{sec:experiments}

\subsection{Experimental Settings}
\paragraph{Baselines.} 
We evaluate MemoSight on two mainstream LLMs, Qwen2.5-7B~\cite{hui2024qwen2} and Llama-3.1-8B~\cite{dubey2024llama}, against six representative baselines categorized into three groups:
(1) CoT Models: The standard instruction model using CoT prompting, alongside the DeepSeek-R1 Distilled model (Distill-R1)~\cite{Guo_2025} 
(2) Trained Reasoning Models: A Vanilla baseline and LightThinker~\cite{zhang2025lightthinker}, both initialized from a Distill-R1 model and further instruction-tuned on the Bespoke-Stratos-17k dataset, a training paradigm that MemoSight also follows.
(3) Training-Free Acceleration: Two post-hoc acceleration methods H2O~\cite{zhang2023ho} and SepLLM~\cite{chen2025sepllm} applied directly to the Vanilla model, which attempt to retain important KV Cache states through heuristic strategies. Details for all baselines are provided in Appendix~\ref{appendix:baseline_details}.

\paragraph{Datasets and Metrics.}
We evaluate on four datasets covering the primary reasoning paradigms of LLMs: mathematical reasoning GSM8K~\cite{cobbe2021training}, algorithmic and compositional reasoning BBH~\cite{suzgun2023challenging}, knowledge-based reasoning MMLU~\cite{hendrycks2020measuring}, and expert-level scientific reasoning GPQA~\cite{rein2024gpqa}. We assess the models from two perspectives: task performance and computational efficiency. Task performance is measured by accuracy (Acc). Computational efficiency is quantified by peak context tokens (Peak) indicating memory footprint and number of tokens generated per second (Speed) reflecting inference throughput.

\paragraph{Implementation Details.} 

All models are trained for 5 epochs with a batch size of 64. We fix the combined budget for reasoning and memory tokens at 4096 across all models.
We set the NTP loss weight $\lambda=0.7$, the compression ratio $c=5$ (i.e., one memory token per five reasoning tokens) and the max foresight offset $d=2$ (i.e., randomly sampling $d' \in \{0,1,2\}$ per training sample, which enables the parallel prediction of three tokens during inference). During evaluation, we employ greedy decoding with a maximum output length of 10240 for all models. We refer readers to Appendix~\ref{appendix:train_details} for more details.

\subsection{Main Results}


Table~\ref{table:exp_main} presents our main results. Consistent with prior findings~\cite{li2025thinking}, Distill-R1 underperforms CoT due to limited instruction-following capabilities. The Vanilla baseline, fine-tuned on reasoning data, exhibits strong task performance. Among the acceleration methods, MemoSight outperforms all the other methods across both the Qwen and Llama model families. For instance, on Qwen, MemoSight exceeds the second-best baseline, LightThinker, by an absolute $3.9$ points in average accuracy. Furthermore, despite its acceleration design, MemoSight achieves average performance comparable to the Vanilla baseline, and even surpasses it on specific datasets (e.g., $+3.54$ on GPQA).

\begin{figure*}[t]
    \centering
    \includegraphics[width=1.0\linewidth]{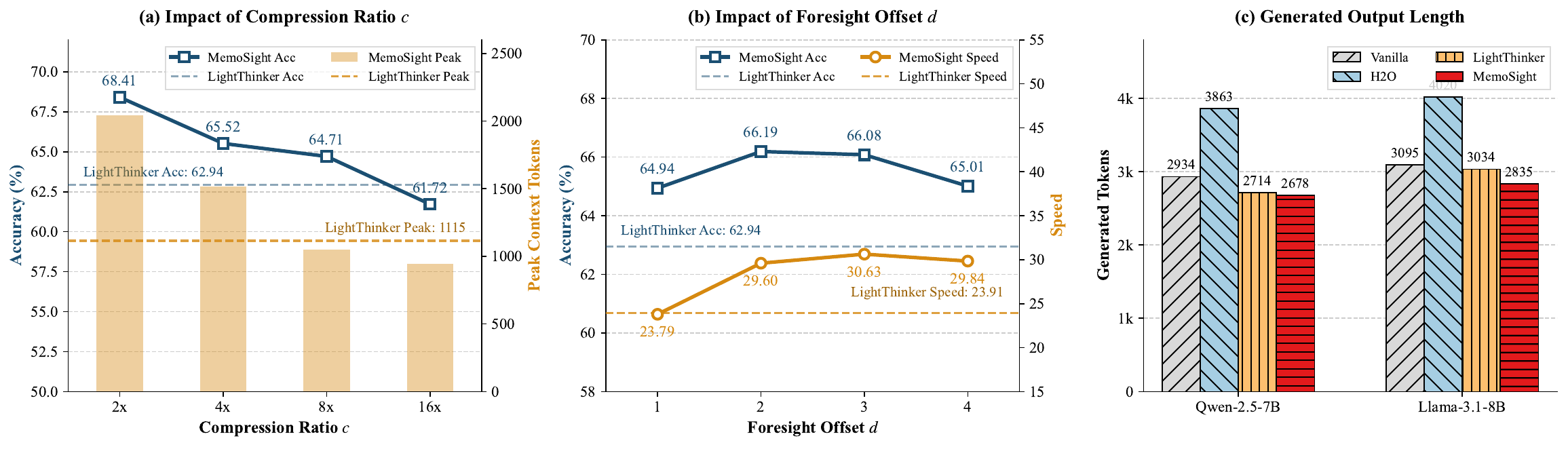}
    \caption{\textbf{Efficiency Analysis.} \textbf{(a) Compression Impact}: Accuracy and peak context token count under compression levels from $2\times$ to $16\times$. Higher compression reduces memory footprint but incurs accuracy degradation. \textbf{(b) Offset Impact}: Accuracy and inference speed with different foresight offsets ($d=1$ to $4$) setting; the same setting is used across training and inference.\textbf{(c) Average generated tokens} across all benchmarks for Vanilla, H2O, LightThinker, and MemoSight on Qwen-2.5-7B and Llama-3.1-8B. MemoSight generates the fewest tokens.}
    \label{fig:analyses}
\vspace{-5pt}
\end{figure*}

In terms of efficiency, training-free acceleration methods fail to improve inference speed over the Vanilla baseline.\footnote{For example, we observe that H2O-generated answers are longer than those of the Vanilla baseline, as shown in Figure~\ref{fig:analyses}(c).} In contrast, MemoSight surpasses both the Vanilla baseline and other acceleration methods, achieving the fastest overall inference. Specifically, MemoSight achieves an average speedup of $23.8\%$ on Qwen and $29.6\%$ on the Llama series compared to the context compression baseline, LightThinker.
Regarding memory consumption, Table~\ref{table:exp_main} shows that at a $5\times$ compression ratio, MemoSight maintains a peak memory footprint comparable to LightThinker and well below the Vanilla baseline. Further analysis in Section~\ref{subsec:efficiency} shows that MemoSight continues to outperform LightThinker even under lower memory budgets (i.e., higher compression ratio).

\section{Analyses}
\label{sec:analyses}

\subsection{Efficiency}
\label{subsec:efficiency}

In this section, we analyze MemoSight's efficiency and its tradeoff with reasoning performance.

\paragraph{How does the compression ratio $c$ affect performance and efficiency?}
Context compression inherently trades off between memory efficiency and reasoning effectiveness. We evaluate MemoSight under compression ratios ranging from $2\times$ to $16\times$. As shown in Figure~\ref{fig:analyses}(a), higher compression ratios consistently reduce memory usage but also degrade accuracy. Compared to LightThinker, MemoSight provides a more favorable tradeoff between efficiency and performance. For example, under $8\times$ compression, MemoSight uses fewer peak tokens than LightThinker while achieving higher average accuracy (+1.77 points). However, aggressive compression at $16\times$ leads to a clear performance drop, suggesting that excessive compression loses critical reasoning information. Table~\ref{table:exp_main} shows that, under a practical $5\times$ compression ratio, MemoSight reduces peak memory usage by 66\% compared to Vanilla decoding with less than a 3\% drop in average accuracy. In Appendix~\ref{appendix:long_efficiency}, we further show that the reduced memory footprint improves inference speed on long sequences.


\paragraph{How does the MTP offset $d$ affect performance and efficiency?}
The parameter $d$ determines how far ahead the MTP module is trained to predict and thus controls the maximum number of parallel predictions enabled during inference. Figure~\ref{fig:analyses}(b) shows the impact of different MTP offsets on accuracy and inference speedup. MemoSight consistently outperforms LightThinker in both metrics across all settings by using MTP.
MemoSight achieves most accuracy improvements at $d=2$ (66.19\%) and remains comparable at $d=3$ (66.08\%); smaller and larger offsets lead to less performance improvements, with $d=1$ achieving 64.94\% and $d=4$ achieving 65.01\%. These results suggest that a moderate look-ahead distance provides the most effective training signal to improve model's reasoning capability. 
Configurations with 2--4 foresight tokens achieve similar inference speedups. Based on these results, we use $d=2$ as the default setting.

\paragraph{Does MemoSight generate more tokens than other methods?}
\citet{zhang2025lightthinker} analyze the generated token lengths of different compression methods and find that LightThinker produces the shortest outputs among existing approaches. We perform the same analysis for MemoSight. Figure~\ref{fig:analyses}(c) reports the average number of generated tokens for Vanilla, H2O, LightThinker, and MemoSight across four datasets. Compared to LightThinker, MemoSight further reduces the generated token length and consistently produces the fewest tokens among all methods. Compared to Vanilla decoding, MemoSight reduces the average number of generated tokens by 9\% on the Qwen model and 8\% on the Llama model.

\subsection{Ablation Study}

\begin{table}[t]
    \centering
    \resizebox{\linewidth}{!}{
        \begin{tabular}{lccccc}
        \toprule
        \textbf{Method} & \textbf{GSM8K} & \textbf{MMLU} & \textbf{GPQA} & \textbf{BBH} & \textbf{Avg.} \\
        \midrule
        \rowcolor{myblue}
        MemoSight           & 89.84 & \textbf{63.49} & \textbf{40.91} & \textbf{73.13} & \textbf{66.84} \\
        \quad w/o UPL & 88.10 & 59.49 & 39.90 & 67.47 & 63.74 \\
        \quad w/o MTP & \textbf{89.99} & 60.86 & 37.37 & 71.72 & 64.99 \\
        \quad w/o AMA & 86.96 & 59.10 & 32.32 & 69.90 & 62.07 \\
        \bottomrule
        \end{tabular}
    }
    \caption{Ablation study of MemoSight's core components, including Uniform Position Layout (UPL), Adaptive Memory Allocation (AMA), and foresight-token-based MTP. Removing any component consistently degrades performance.}
    \label{table:ablation}
\vspace{-5pt}
\end{table}

We further ablate the core components of MemoSight, including Uniform Position Layout (UPL) which assigns uniformly interpolated position IDs to memory tokens over the compressed span, Adaptive Memory Allocation (AMA) which scales the number of memory tokens according to the reasoning step length; and foresight-token-based MTP. Detailed ablation settings are provided in Appendix~\ref{appendix:ablation_details}. We evaluate all variants on Qwen and report the results in Table~\ref{table:ablation}. 

The full model achieves the highest average accuracy (66.84), while removing any component consistently degrades performance. Among the components, AMA is particularly important for the knowledge-intensive GPQA benchmark, where removing AMA causes a large accuracy drop (-8.59). In contrast, removing UPL leads to the largest degradation on BBH (-5.66). Finally, removing MTP decreases the average accuracy by 1.85, indicating that the foresight-token mechanism improves reasoning performance in addition to accelerating inference.

\begin{table}[t]
    \centering
    \resizebox{\linewidth}{!}{
        \begin{tabular}{lcccccc}
        \toprule
        \textbf{Method} & \textbf{Param.} & \textbf{GSM8K} & \textbf{MMLU} & \textbf{GPQA} & \textbf{BBH} & \textbf{Avg.} \\
        \midrule
        \rowcolor{myblue}
        MemoSight     & \textbf{4K} & \textbf{89.84} & \textbf{63.49} & \textbf{40.91} & \textbf{73.13} & \textbf{66.84} \\
        MTP Middle & 103M & 88.17 & 61.25 & 38.89 & 69.49 & 64.45 \\
        MTP Last   & 103M & 85.06 & 61.34 & 32.83 & 71.11 & 62.59 \\
        \bottomrule
        \end{tabular}
    }
    \caption{Comparison between MemoSight and traditional MTP variants. Traditional MTP modules introduce more parameters but achieve lower accuracy.}
    \label{table:mtp_variants}
\vspace{-5pt}
\end{table}

\subsection{Comparison with Traditional MTP}
\label{subsec:MTP}

We vary the MTP mechanisms used in MemoSight to study their effect on accuracy. Specifically, we integrate DeepSeek-V3's MTP variant (MTP Last)~\cite{liu2024deepseek} and a middle-layer variant (MTP Middle)~\cite{wang2025mtp} into MemoSight for comparison. Detailed descriptions of these variants are provided in Appendix~\ref{appendix:traditional_mtp}.
Table~\ref{table:mtp_variants} suggests that combining MTP with context compression is non-trivial. Both traditional MTP variants degrade performance compared to the MTP-free baseline (i.e., MemoSight w/o MTP in Table~\ref{table:ablation}). We attribute this behavior to the large parameter overhead of traditional MTP modules (100M+ parameters), which makes joint optimization with context compression more difficult.
In contrast, MemoSight unifies MTP and compression within a special-token-based framework, enabling joint optimization with only 4K additional parameters.



\section{Related Work}
To tackle CoT computational efficiency issues, recent studies CoConut~\cite{hao2025traininglargelanguagemodels} and its variants CODI~\cite{shen2025codicompressingchainofthoughtcontinuous} and SIM-CoT~\cite{wei2025simcotsupervisedimplicitchainofthought} propose latent reasoning that performs reasoning in the high dimensional latent space. However, latent reasoning lacks interpretability and is prone to learning dataset artifacts in its reasoning ~\cite{zhang2025latent}. As an alternative, LightThinker~\cite{zhang2025lightthinker} explicitly compresses CoT into its discrete memory tokens step-by-step. MemoSight follows LighThinker in its design but further equips its memory tokens with a special position layout inspired by~\cite{zhao2024position} that enhances its compression capability a priori.

To the best of our knowledge, we are the first to combine MTP with CoT compression for CoT inference acceleration. We empirically show that such combination is non-trivial as most existing MTP frameworks require architectural modifications to the LLM~\cite{liu2024deepseek, gloeckle2024better}, which hinder performance when combined with CoT compression. Instead, in MemoSight, we propose to integrate MTP via a unified way through special tokens and a tailored position layout following~\citep{gerontopoulos2025multi} --- the same mechanism that we integrate CoT compression and demonstrate not only accelerated inference but also improved performance. We provide more comprehensive discussion of related work in Appendix~\ref{appendix:related_work}.

\section{Conclusion}


We present MemoSight to address memory and latency bottlenecks in Chain-of-Thought reasoning. While naively combining context compression and multi-token prediction degrades reasoning performance, MemoSight overcomes this through a unified design using memory tokens and foresight tokens, both of which equipped with a dedicated position layout. Compared to the CoT fine-tuned LLM counterpart, MemoSight reduces the KV cache footprint by 66\% and improves inference speed by 56\% with less than 3\% answer accuracy degradation, reaching a new state-of-the-art efficiency-performance balance.

\section*{Limitations}
Despite its effectiveness, MemoSight has several limitations both on CoT compression and MTP. 

Compared to CoT prompting, MemoSight enables longer CoT reasoning by effectively shortening each reasoning step. However, we remark that MemoSight does not enable infinite long reasoning, as memory tokens still accumulate within the KV cache during each CoT step. We believe techniques beyond compression, such as memory management~\cite{zhu2026lightthinker++} is required to achieve infinite long reasoning. MemoSight experiences performance degradation when pushed to high compression ratios where $16\times$ compression ratio already significantly degrades average accuracy in our experiments.

We integrate MTP into our unified framework and achieve both performance improvement and inference acceleration. However, we notice that our inference acceleration is only 24\% for Qwen and 30\% for Llama. Our analysis shows that even by training on more parallel predictions with larger $d$, the acceleration does not further improve as more futuristic tokens are mostly rejected through verification. It remains question how to better train MTP to learn more future token predictions in parallel reliably.

\bibliography{custom}

\newpage
\appendix





\section{Inference Procedure}
\label{appendix:infer_procedure}

We outline the iterative inference procedure in Algorithm~\ref{alg:MemoSight_simplified}.

\begin{algorithm}[ht]
\small
\caption{MemoSight Iterative Inference}
\label{alg:MemoSight_simplified}
\begin{algorithmic}[1]

\REQUIRE Prompt $P$, model $\mathcal{M}$ with KV cache $S$, foresight offset $d$, compression rate $c$, special tokens $\langle \text{e} \rangle$ (segment boundary) and $\langle \text{eos} \rangle$, draft tokens $\hat{R}$, and accepted tokens $A$.

\STATE $S \leftarrow \text{Prefill}(P)$ \COMMENT{Initialize KV cache}
\WHILE{$\langle \text{eos} \rangle \notin A$}
    \STATE $R_{\text{current}} \leftarrow \emptyset$ \COMMENT{Initialize segment tokens}
    \STATE \textbf{// Phase 1: Foresight-based Acceleration}
    \REPEAT
        \STATE $A \leftarrow \emptyset$ \COMMENT{Initialize accept tokens}
        \STATE $\hat{R}, A \leftarrow \text{SpeculativeStep}(\mathcal{M}, S, d)$
        \STATE $S \leftarrow S \oplus A$ \COMMENT{Append to cache}
        \STATE $R_{\text{current}} \leftarrow R_{\text{current}} \oplus A$ \COMMENT{Accumulate segment}
    \UNTIL{$\langle \text{e} \rangle \in A$ \OR $\langle \text{eos} \rangle \in A$}
    
    \STATE \textbf{// Phase 2: Dynamic Compression}
    \IF{$\langle \text{e} \rangle \in A$}
        \STATE $l \leftarrow \lceil |R_{\text{current}}| / c \rceil$ 
        \STATE $M \leftarrow \text{Forward}(\mathcal{M}, \langle \text{m}_1, \dots, \text{m}_l, \text{o} \rangle)$ 
        \STATE $S \leftarrow \text{UpdateCache}(S, R_{\text{current}}, M)$ \COMMENT{Evict verbose and insert memory}
    \ENDIF
\ENDWHILE
\end{algorithmic}
\end{algorithm}

\section{Experimental Details}

\subsection{Baseline Details}
\label{appendix:baseline_details}

During evaluation, we employ greedy decoding with a maximum output length of 10,240 tokens for all models. We compare MemoSight against the following baselines:

\begin{itemize}[topsep=4pt, itemsep=4pt, parsep=2pt, leftmargin=*]
    \item \textbf{CoT}: A baseline that applies few-shot Chain-of-Thought (CoT) prompting to the Qwen2.5-7B~\cite{hui2024qwen2} and Llama-3.1-8B~\cite{dubey2024llama} models without additional training.

    \item \textbf{Distill-R1}~\cite{Guo_2025}: A reasoning model distilled from DeepSeek-R1's response data, built upon the Qwen and Llama architectures.
    
    \item \textbf{Vanilla}: A standard full-parameter instruction-tuned model. Operating without any compression or acceleration mechanisms, it serves as the empirical upper bound for reasoning accuracy.

    \item \textbf{H2O}~\cite{zhang2023ho}: A training-free KV cache eviction strategy that retains "Heavy Hitter" tokens and a local window. We apply H2O to the Vanilla model using a sliding window of 1024 and a heavy-hitter budget of 512 tokens.

    \item \textbf{SepLLM}~\cite{chen2025sepllm}: A training-free framework that preserves KV caches for initial tokens, separators, and a local window. We configure SepLLM with an initial cache size of 384, a separator budget of 64, and a local window of 256, maintaining a total cache capacity of 1024.

    \item \textbf{LightThinker}~\cite{zhang2025lightthinker}: A post-training method that compresses each reasoning step into a fixed number of memory tokens. Upon reaching a step boundary, the model generates memory tokens to summarize the context, after which the original reasoning tokens are evicted from the KV cache.
\end{itemize}

\subsection{Training Details}
\label{appendix:train_details}

The Vanilla baseline, LightThinker and MemoSight are initialized from DeepSeek-R1-Distill~\cite{Guo_2025} and trained on the Bespoke-Stratos-17k (BS17K) dataset for 5 epochs. Experiments are conducted on 8 H200 GPUs using DeepSpeed ZeRO-3 offload. We use a micro-batch size of 2 and 4 gradient accumulation steps, yielding a global batch size of 64. We employ a cosine learning rate schedule with a 0.05 warmup ratio. The peak learning rate for Vanilla is set to 1e-5, while for the other models, it is set to 2e-5.

All the CoT token length (including memory tokens) has been set to 4096; to accommodate the additional foresight tokens introduced during training, MemoSight is extended to a maximum sequence length of 8192. For MemoSight specifically, we set the compression ratio to $c=5$ and randomly sample the foresight offset $d' \in \{0, 1, 2\}$ during data construction. The standard LM loss and the MTP loss are weighted at 0.7 and 0.3, respectively (i.e., $\lambda = 0.7$).

\subsection{Ablation Details}
\label{appendix:ablation_details}

This section describes the experimental setup for our ablation studies, which evaluate the contributions of Uniform Position Layout (UPL), Adaptive Memory Allocation (AMA), and foresight-token-based multi-token prediction (MTP).

\begin{itemize}[topsep=4pt, itemsep=4pt, parsep=2pt, leftmargin=*]

\item \textbf{MemoSight w/o UPL:}
This variant assigns monotonically increasing position IDs to memory tokens, rather than uniformly interpolated PIDs. Foresight tokens are assigned PIDs following the standard offset rule (i.e., $t+d-1$ when predicting the token at $t+d$).

\item \textbf{MemoSight w/o AMA:}
Instead of using adaptive memory allocation with a fixed compression ratio, this variant allocates a fixed budget of 9 memory tokens per reasoning step, following the LightThinker configuration.

\item \textbf{MemoSight w/o MTP:}
This variant removes foresight tokens and the multi-token prediction objective, relying solely on memory tokens for context compression.

\end{itemize}

All ablation experiments are conducted on the Qwen series models. Training settings are kept identical across all variants and follow the same configuration as MemoSight and LightThinker.

\subsection{Traditional MTP Details}
\label{appendix:traditional_mtp}

To assess the effectiveness of our foresight-token-based MTP, we establish traditional MTP baselines operating under identical context compression settings. Following DeepSeek-V3~\cite{liu2024deepseek}, the standard configuration sequentially appends two transformer blocks to the final layer's hidden states, predicting the next two tokens via a shared LM head. Furthermore, inspired by \citet{wang2025mtp}, who demonstrated that leveraging intermediate representations yields superior performance in speech tasks, we evaluate an additional variant that feeds intermediate layer hidden states—rather than the final layer's output—into the MTP module.

\subsection{Evaluation Details}

During inference, we evaluate all models using greedy decoding with a repetition penalty of 1.1. Prompt configurations for Table~\ref{table:exp_main} are as follows: Vanilla, H2O, SepLLM, LightThinker, and MemoSight share the same system prompt (Figure~\ref{fig:system_prompt_oth}) and task prompts (Figure~\ref{fig:task_prompt_oth}). Distill-R1 uses these task prompts but omits the system prompt. The CoT baseline uses a few-shot system prompt (Figure~\ref{fig:system_prompt_cot}) alongside task-specific prompts for different benchmarks (Figure~\ref{fig:task_prompt_cot}). For MMLU~\cite{hendrycks2020measuring} and GPQA~\cite{rein2024gpqa}, multiple-choice options are randomized to prevent positional bias.

\section{More Analyses and Discussions}
\label{appendix:more_analyses}

\subsection{Long-Context Efficiency Analysis}
\label{appendix:long_efficiency}
\begin{figure*}[htbp]
    \centering
    \includegraphics[width=1.0\linewidth]{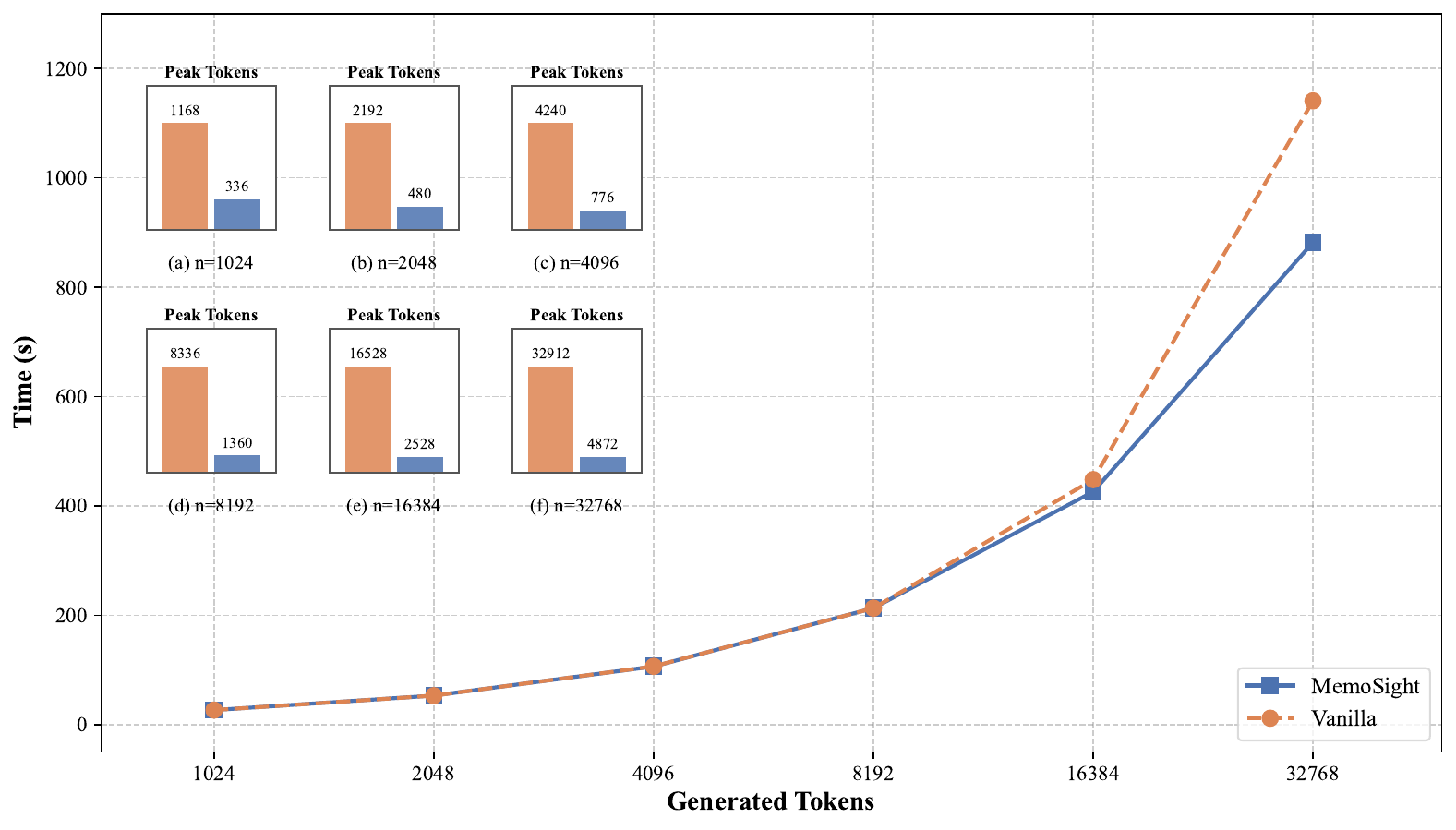}
    \caption{Time and memory efficiency evaluation. The main plot shows the inference time of MemoSight and Vanilla as the number of generated tokens increases. The inset charts (a)-(f) compare the peak token usage at different sequence lengths.}
    \label{fig:length_analysis}
    \vspace{-5pt}
\end{figure*}

Figure~\ref{fig:length_analysis} compares the peak memory and inference time of MemoSight against the vanilla baseline during autoregressive generation. We set a fixed compression ratio of 7, compressing every 56 tokens. Results show that context compression effectively reduces memory consumption, lowering peak memory by over 80\%. Regarding inference speed, since KV cache reduction primarily accelerates the attention layer rather than other modules, the speed improvement only becomes evident when the context length exceeds 8k.

\subsection{Case Study}
\begin{figure*}[htbp]
    \centering
    \includegraphics[width=1.0\linewidth]{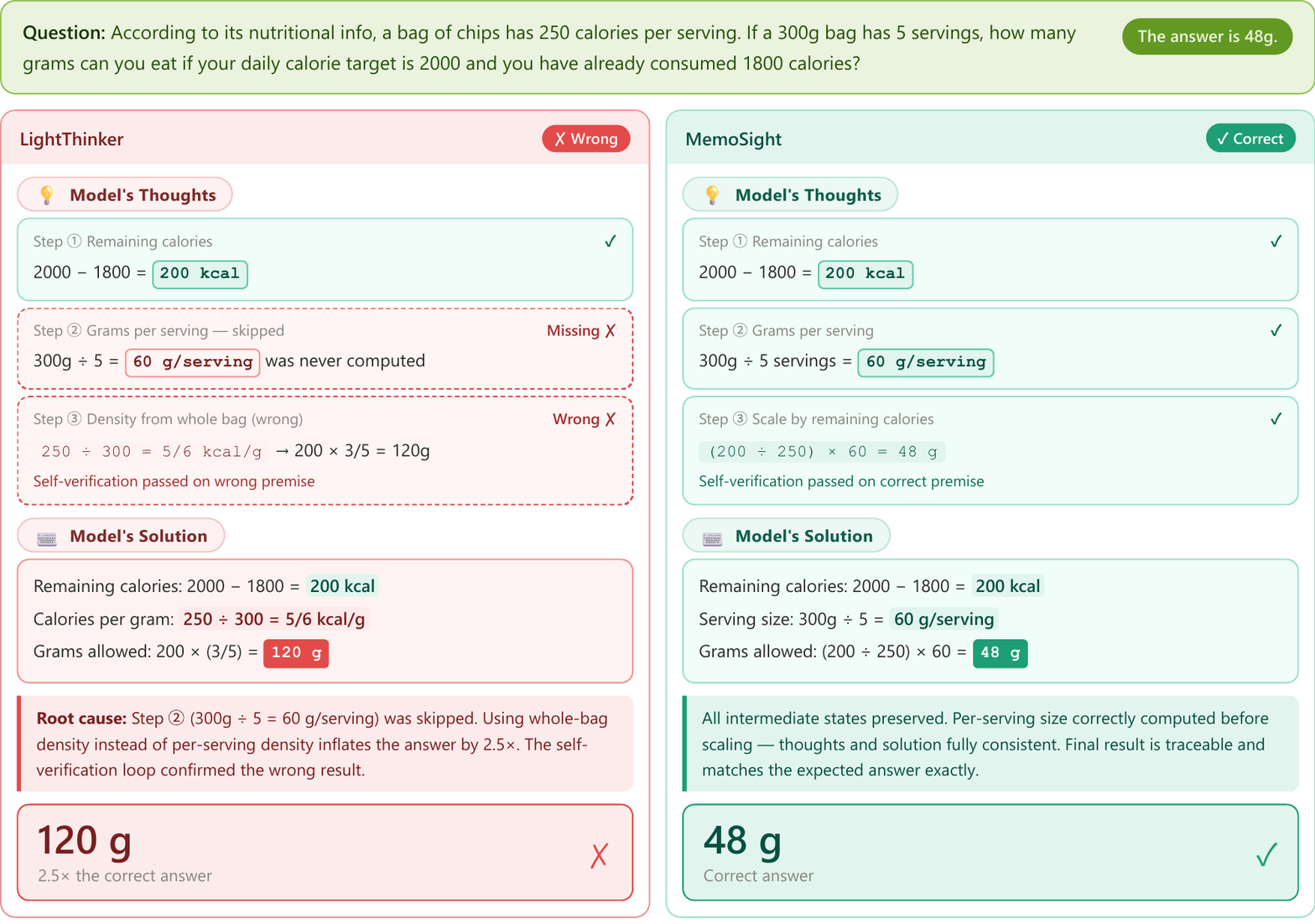}
    \caption{Case Study comparing the reasoning trajectories of LightThinker and MemoSight.}
    \label{fig:case_study}
\vspace{-5pt}
\end{figure*}

Figure~\ref{fig:case_study} illustrates how preserving intermediate reasoning states prevents cascading errors in multi-step calculations. In this nutritional calculation problem, the baseline model, LightThinker, correctly computes the remaining calorie budget but skips calculating the grams per serving. As a result, it computes an incorrect energy density by dividing the per-serving calories by the whole-bag weight ($250 \div 300$), leading to a flawed self-verification and an incorrect final answer of 120 g. In contrast, MemoSight explicitly preserves all intermediate states. It calculates the correct serving size ($300 \div 5 = 60$) before applying the calorie constraint. This ensures the final scaling step is grounded in the correct premise ($(200 \div 250) \times 60$), allowing the model to consistently arrive at the correct answer of 48 g.

\subsection{Impact of Loss Weights}
\begin{figure*}[htbp]
    \centering
    \includegraphics[width=1.0\linewidth]{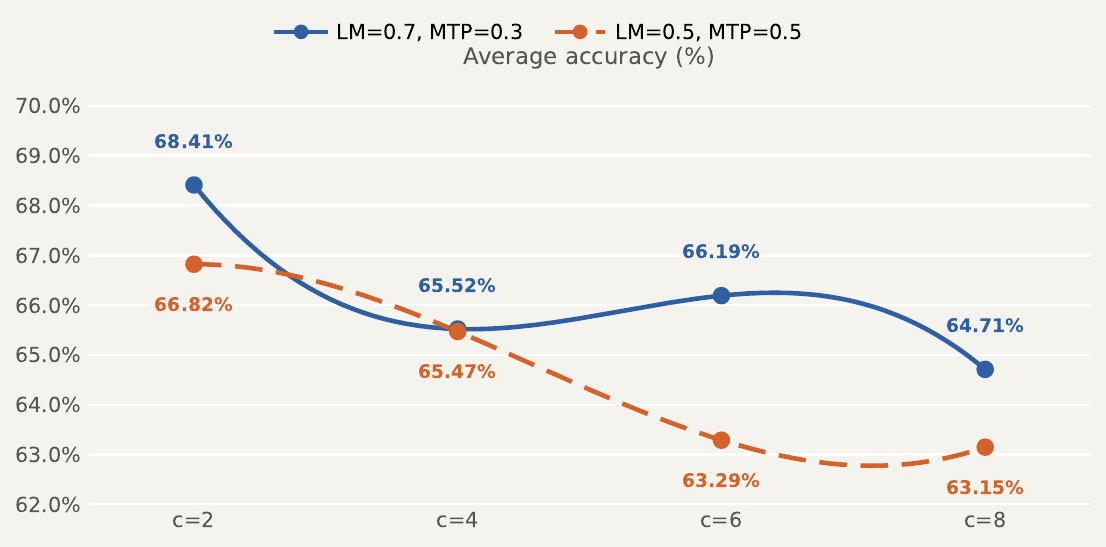}
    \caption{Impact of loss weight configuration ($\lambda$) on average accuracy across varying compression ratios ($c$). The blue solid line represents a higher weight on the standard LM loss ($\lambda=0.7$), while the orange dashed line represents an equal weighting ($\lambda=0.5$).}
    \label{fig:loss_weight_smooth}
    \vspace{-5pt}
\end{figure*}

We investigate the effect of the loss weight distribution by comparing two configurations: a higher weight on the standard Language Modeling (LM) loss $(\lambda=0.7)$ and an equal weighting $(\lambda=0.5)$. As illustrated in Figure~\ref{fig:loss_weight_smooth}, the $0.7/0.3$ configuration consistently outperforms the equal weighting baseline across various compression ratios ($c$). Notably, allocating a larger weight to the MTP objective ($0.5$) degrades average accuracy, suggesting that excessive planning supervision interferes with the primary generation task. Conversely, a moderate MTP weight ($0.3$) achieves a more effective balance, preserving core reasoning capabilities while still benefiting from the lookahead training signal.

\section{Related Work}
\label{appendix:related_work}

\subsection{Context Compression} 

Context compression mitigates the computational and memory overhead of long-sequence processing by condensing contexts while preserving semantic information. Since our work focuses on long-form generation rather than prompt prefilling, we exclude methods that primarily compress the input context during the prefilling stage, such as AutoCompressor~\cite{chevalier2023adapting}, ICAE~\cite{ge2023context}, LLMLingua~\cite{jiang2023llmlingua}, 500xCompressor~\cite{li2025500xcompressor}, ActivationBeacon~\cite{zhang2024long}, PyramidKV~\cite{cai2024pyramidkv}, Perception Compressor~\cite{tang2025perception}, SnapKV~\cite{li2024snapkv}, GMSA~\cite{tang2025gmsa}, PoC~\cite{zhao2026poc}, COMI~\cite{tang2026comi}, RAM~\cite{tang2026read}, SAC~\cite{liu2025autoencoding}, and CoMeT~\cite{zhao2026comet}.

Generation-phase context compression methods generally fall into three paradigms: latent reasoning, explicit token selection, and implicit latent condensation.

\noindent \textbf{1) Latent reasoning.}
To avoid the verbosity of explicit chain-of-thought (CoT) reasoning, methods such as Coconut~\cite{hao2025traininglargelanguagemodels} and SoftCoT~\cite{xu2025softcot} perform reasoning in continuous latent spaces. To improve training stability and mitigate latent collapse, CODI~\cite{shen2025codicompressingchainofthoughtcontinuous} aligns latent states with natural-language CoT trajectories, while SIM-CoT~\cite{wei2025simcotsupervisedimplicitchainofthought} introduces auxiliary step-level supervision. However, \citet{zhang2025latent} argue that current latent reasoning models often exhibit pseudo-reasoning behaviors rather than genuine reasoning capabilities.

\noindent \textbf{2) Explicit token selection.}
Early approaches remove tokens according to importance metrics, such as LLMLingua~\cite{jiang2023llmlingua, pan2024llmlingua}. However, aggressive token pruning may damage local coherence and logical dependencies. More recent methods incorporate structural information during compression. For example, Context Compressor~\cite{zhou2025context} leverages discourse trees to preserve global structure, SWEzze~\cite{jia2026compressing} extracts minimal sufficient subsequences for code repositories, and TokenSkip~\cite{xia2025tokenskip} dynamically omits redundant reasoning tokens during decoding.

\noindent \textbf{3) Implicit latent condensation.}
This paradigm compresses contexts into continuous latent embeddings, commonly represented as memory tokens. AnLLM~\cite{pang2024anchor} employs anchor-based self-attention to distill sequence information into specialized anchor tokens. LightThinker~\cite{zhang2025lightthinker} compresses intermediate reasoning traces into compact gist tokens and discards verbose reasoning states to reduce KV cache growth during generation. Building upon this idea, LightThinker++~\cite{zhu2026lightthinker++} introduces adaptive memory management through memory primitives and trajectory synthesis, enabling stable KV cache usage in long-horizon reasoning tasks. Unlike prefilling-based compression methods, these approaches explicitly target the incremental memory overhead arising during autoregressive generation.

\subsection{Multi-Token Prediction}

Multi-token prediction (MTP) extends the standard next-token prediction objective by simultaneously predicting multiple future tokens, thereby densifying training signals and encouraging longer-horizon planning. Early work such as ProphetNet~\cite{qi2020prophetnet} introduced MTP for sequence-to-sequence learning, but its multi-stream attention mechanism scales poorly to large language models.

Recent studies address this limitation using auxiliary prediction heads. \citet{gloeckle2024better} employ parallel decoding heads to improve generative modeling, while \citet{liu2024deepseek} introduce sequential prediction heads to enhance implicit planning within hidden states. However, these gains are mainly observed during large-scale pretraining. To better support fine-tuning scenarios, \citet{gerontopoulos2025multi} recently proposed a special-token-based MTP formulation.

To the best of our knowledge, our work is the first to integrate multi-token prediction with context compression, further accelerating reasoning through speculative decoding~\cite{cai2024medusa,li2024eagle}.

\begin{figure*}[htbp]
    \centering
    \begin{tcolorbox}[
        colback=yellow!8,      
        colframe=black!70,  
        arc=2mm,            
        boxrule=1.0pt,      
        left=4pt, right=4pt, top=4pt, bottom=4pt 
    ]
    \vspace{0.2cm}
    \textbf{System Prompt:} \\
    Below is a question. Please think through it step by step, and then provide the final answer. If options are provided, please select the correct one.
    \vspace{0.3cm}
    
    \#\# Output format: \\
    Use ``\textless THOUGHT\textgreater...\textless/THOUGHT\textgreater'' to outline your reasoning process, and enclose the final answer in `\textbackslash boxed\{\}'.
    \vspace{0.3cm}
    
    \#\# Example 1: \\
    Question: \\
    What is 2 + 3? \\
    Output: \\
    \textless THOUGHT\textgreater First, I recognize that this is a simple addition problem. Adding 2 and 3 together gives 5.\textless/THOUGHT\textgreater \\
    Therefore, the final answer is \textbackslash boxed\{5\}.
    \vspace{0.3cm}
    
    \#\# Example 2: \\
    Question: \\
    What is 2 + 3? \\
    A. 4 \\
    B. 5 \\
    C. 10 \\
    
    Output: \\
    \textless THOUGHT\textgreater First, I recognize that this is a simple addition problem. Adding 2 and 3 together gives 5.\textless/THOUGHT\textgreater \\
    Therefore, the final answer is \textbackslash boxed\{B\}.
    \vspace{0.2cm}
    \end{tcolorbox}
    \caption{System prompt for Qwen2.5-7B-Instruct and Llama3.1-8B-Instruct.}
    \label{fig:system_prompt_cot}
\end{figure*}

\begin{figure*}[htbp]
    \centering
    \begin{tcolorbox}[
        colback=yellow!8,      
        colframe=black!70,     
        arc=2mm,               
        boxrule=1.0pt,         
        left=4pt, right=4pt, top=4pt, bottom=4pt 
    ]
    \vspace{0.2cm}
    \textbf{GSM8K:} \\
    Return your final response within \textbackslash boxed\{\}. \textcolor{purple}{\{Question\}}
    \vspace{0.4cm}
    
    \textbf{MMLU:} \\
    Please select the option that best answers the question. Return your final response within \textbackslash boxed\{\}. Here are the Question: \\
    \textcolor{purple}{\{Question\}}
    \vspace{0.4cm}
    
    \textbf{GPQA:} \\
    Given a question, please select the option that best answers it. Return your final response within \textbackslash boxed\{\}. \textcolor{purple}{\{Question\}}
    \vspace{0.4cm}
    
    \textbf{BBH:} \\
    Return your final response within \textbackslash boxed\{\}. If options are provided, please select the correct one. \\
    \textcolor{purple}{\{Question\}}
    \vspace{0.2cm}
    \end{tcolorbox}
    \caption{Task prompt for Qwen2.5-7B-Instruct and Llama3.1-8B-Instruct.}
    \label{fig:task_prompt_cot}
\end{figure*}

\begin{figure*}[htbp]
    \centering
    \begin{tcolorbox}[
        colback=yellow!8,      
        colframe=black!70,     
        arc=2mm,               
        boxrule=1.0pt,         
        left=4pt, right=4pt, top=4pt, bottom=4pt 
    ]
    \vspace{0.2cm}
    \textbf{System Prompt:} \\
    Your role as an assistant involves thoroughly exploring questions through a systematic long thinking process before providing the final precise and accurate solutions. This requires engaging in a comprehensive cycle of analysis, summarizing, exploration, reassessment, reflection, backtracking, and iteration to develop well-considered thinking process. Please structure your response into two main sections: Thought and Solution. In the Thought section, detail your reasoning process using the specified format: \textless|begin\_of\_thought|\textgreater\ \{thought with steps separated with `\textbackslash n\textbackslash n'\} \textless|end\_of\_thought|\textgreater\ Each step should include detailed considerations such as analyzing questions, summarizing relevant findings, brainstorming new ideas, verifying the accuracy of the current steps, refining any errors, and revisiting previous steps. In the Solution section, based on various attempts, explorations, and reflections from the Thought section, systematically present the final solution that you deem correct. The solution should remain a logical, accurate, concise expression style and detail necessary steps needed to reach the conclusion, formatted as follows: \textless|begin\_of\_solution|\textgreater\ \{final formatted, precise, and clear solution\} \textless|end\_of\_solution|\textgreater\ Now, try to solve the following question through the above guidelines:
    \vspace{0.2cm}
    \end{tcolorbox}
    \caption{The shared system prompt applied to Vanilla, H2O, SepLLM, LightThinker, and MemoSight across the Qwen and Llama series.}
    \label{fig:system_prompt_oth}
\end{figure*}

\begin{figure*}[htbp]
    \centering
    \begin{tcolorbox}[
        colback=yellow!8,      
        colframe=black!70,     
        arc=2mm,               
        boxrule=1.0pt,         
        left=4pt, right=4pt, top=4pt, bottom=4pt 
    ]
    \vspace{0.2cm}
    \textbf{GSM8K/MMLU/GPQA/BBH:} \\
    Return your final response within \textbackslash boxed\{\}. \textcolor{purple}{\{Question\}}
    \vspace{0.2cm}
    \end{tcolorbox}
    \caption{The shared task prompt applied to Vanilla, H2O, SepLLM, LightThinker, and MemoSight across the Qwen and Llama series.}
    \label{fig:task_prompt_oth}
\end{figure*}

\end{document}